\documentclass{article}

    \usepackage[preprint]{tackling_climate_workshop_style}

\usepackage[utf8]{inputenc} 
\usepackage[T1]{fontenc}    
\usepackage{hyperref}       
\usepackage{url}            
\usepackage{booktabs}       
\usepackage{amsfonts}       
\usepackage{nicefrac}       
\usepackage{microtype}      
\usepackage{graphicx}
\usepackage{amsmath}

\title{\textit{Generalizable AI Model for Indoor Temperature Forecasting Across Sub-Saharan Africa}}

\author{%
  Zainab Akhtar \\
  Open Development \& Education\\
  \texttt{zainab@opendeved.net} \\
  \And
  Eunice Jengo\\
  Open Development \& Education \\
  \texttt{eunice@opendeved.net} \\
  \AND
  Björn Haßler\\
  Open Development \& Education \\
  \texttt{bjoern@opendeved.net} \\
}

\begin{document}

\maketitle

\begin{abstract}
  This study presents a lightweight, domain-informed AI model for predicting indoor temperatures in naturally ventilated schools and homes in Sub-Saharan Africa. The model extends the Temp-AI-Estimator framework, trained on Tanzanian school data, and evaluated on Nigerian schools and Gambian homes. It achieves robust cross-country performance using only minimal accessible inputs, with mean absolute errors of 1.45°C for Nigerian schools and 0.65°C for Gambian homes. These findings highlight AI's potential for thermal comfort management in resource-constrained environments.
\end{abstract}

\section{Introduction}
Thermal comfort is essential for health, cognition, learning and well-being, yet most schools and homes in Sub-Saharan Africa (SSA) depend on natural ventilation without climate control, exposing occupants to extreme indoor heat that undermines concentration and academic performance [12, 13]. Despite its importance, thermal comfort remains poorly monitored in these settings, limiting data-driven infrastructure planning. While indoor temperature provides a practical proxy for comfort, existing AI prediction models assume sensor-rich environments with stable electricity and dense data, conditions rarely present in SSA [17, 5].

This creates a critical gap for low-resource contexts with limited infrastructure, heterogeneous buildings, and sparse data. A lightweight model relying only on outdoor weather and simple building characteristics is needed to predict thermal discomfort.

We develop a scalable AI model by adapting the Temp-AI-Estimator framework [3] for SSA contexts. Key contributions include: 1) a generalizable approach for low-resource buildings, and 2) cross-country validation spanning Tanzania, Nigeria, and The Gambia. By centering thermal comfort in educational and domestic settings, this work advances environmental data science while supporting educational equity, climate adaptation, and evidence-based policymaking in vulnerable regions.

\section{Related Work}
Indoor temperature prediction has been extensively studied using machine learning approaches, though most work focuses on high-income regions. Decision-tree methods (Random Forest, XGBoost, LightGBM) achieve mean errors below 1°C with outdoor climate variables and building descriptors [8, 1], but struggle to generalize across climatic contexts. Deep learning techniques like hybrid ANN-LSTM models [7] and error-corrected LSTMs [16] excel at modeling temporal dynamics but require dense, high-frequency data rarely available in low-resource settings. While ensemble strategies have improved cross-building generalization [17], applications in SSA contexts remain scarce.

Recent research emphasizes generalizability and resource efficiency. Sophisticated models like MITP-Net [15] and CONST-LSTM [14] integrate spatio-temporal consistency but depend on comprehensive sensor data and high computation, limiting rural deployment. Emerging causal modeling approaches show promise, with Mun and Park [10] demonstrating that double machine learning outperforms traditional methods by identifying causal relationships.

This study bridges this gap by developing a generalizable forecasting model for low-resource, naturally ventilated buildings using harmonized datasets from Tanzania, The Gambia, and Nigeria, a unique cross-country integration. 

\section{Data}

\textit{Indoor Temperature Data}

Three datasets from various geographic locations, covering different years, seasons, and building types were selected for representativeness. \textbf{1) Tanzania:} Data from 41 primary schools \footnote{Sensor data from 45 Tanzanian schools were collected but after preprocessing, continuous data from 41 schools remained [9]. All school locations are visualized on this \href{https://huggingface.co/spaces/zakhtar/TanzanianSchools_WebMap}{Web Application}} located across three provinces (Mara, Pwani, and Dodoma). The data spans approximately four months (17 July – 4 December 2023), covering both dry and rainy seasons. \textbf{2) The Gambia:} Data from two experimental rural houses located at Medical Research Councils Unit The Gambia's, field station at \href{https://www.google.com/maps/place/Medical+Research+Council+Gambia/@13.5737393,-14.9250072,254m/data=!3m1!1e3!4m6!3m5!1s0xeef310ecd763463:0xe2bbea62d79d91d4!8m2!3d13.5734232!4d-14.9249861!16s%2Fg%2F11q969k4mv?entry=ttu&g_ep=EgoyMDI1MDcyMy4wIKXMDSoASAFQAw%3D%3D}{Wali Kunda} were collected over a five-week period, from 5 October -- 29 September 2017 [4]. These houses differed primarily in roof color, one with a white reflective roof, and another with a traditional darker roof. \textbf{3) Nigeria:} One public primary school, \href{https://www.google.com/maps/place/Central+School+Umuduru/@5.6909509,7.2409364,15z/data=!4m10!1m2!2m1!1s+CENTRAL+SCHOOL+UMUDURU+in+Imo,+Okigwe,+Nigeria!3m6!1s0x104303ffaa939f3f:0x53bc1168bca04b0f!8m2!3d5.6908939!4d7.2410096!15sCi5DRU5UUkFMIFNDSE9PTCBVTVVEVVJVIGluIEltbywgT2tpZ3dlLCBOaWdlcmlhkgERZWxlbWVudGFyeV9zY2hvb2yqAXEQASoaIhZjZW50cmFsIHNjaG9vbCB1bXVkdXJ1KAAyHxABIhtt-kqz2ccT7VzGPRHrHMBhSayGw7LiJ7E6WGQyMBACIixjZW50cmFsIHNjaG9vbCB1bXVkdXJ1IGluIGltbyBva2lnd2UgbmlnZXJpYeABAA!16s%2Fg%2F11j1c1p3gs?entry=ttu&g_ep=EgoyMDI1MDcyMy4wIKXMDSoASAFQAw%3D%3D}{Central School Umuduru}, in Imo State was selected from a broader set of six schools [11]. For the selected school, indoor temperature data was collected during the rainy season, for approximately 1 month (5 – 28 May 2018).

\textit{Outdoor Weather Data}

To complement indoor data, external weather variables were integrated from two reliable global weather datasets: \textbf{1) Open-Meteo:} Hourly weather data provided through a globally accessible API, offering accurate and timely environmental information for each study location [18], and \textbf{2) ERA5 Reanalysis:} Comprehensive hourly weather data, providing long-term consistent meteorological data, critical for accurately capturing external environmental influences on indoor conditions [6].

\textit{Variables and Feature Set}

Appendix~\ref{appendix:methodology} illustrates the comprehensive set of features employed for the predictive AI model, grouped into four distinct categories. The \textbf{Target Variable} is indoor temperature, directly measured within buildings where the \textbf{Temporal Features} were derived from timestamps. \textbf{Contextual Features} include classroom area, occupancy, roof color, and ceiling board presence. Roof color was categorized as light (white), medium (silver with slight rust), and dark (red/brown with significant rust). \textbf{External Weather Features} consist of three variables obtained from Open-Meteo (air temperature, relative humidity, and dew point), two from ERA5 (surface pressure and total precipitation), alongside sun azimuth and sun altitude calculated using the 'pvlib' Python library [2]. This feature set was carefully curated based on existing literature, domain expertise, and by analyzing feature importance via SHAP plots, to ensure robust, meaningful, and interpretable predictions of indoor temperature.

\section{Methodology}
The methodology adapts and extends the Temp-AI-Estimator framework [3], training on the longer Tanzanian dataset ($\sim$4 months) and evaluating on shorter Nigeria and The Gambia datasets ($\sim$1 month). Minute-level data is aggregated to hourly intervals and structured using 12-hour sliding windows for 24-hour temperature forecasts. The adapted architecture integrates an LSTM backbone for temporal dependencies, external-correction and physical-modulation branches for building-specific adjustments, and a domain discriminator with adversarial training to promote cross-country generalization. Training employed supervised regression loss, adversarial domain loss, and Adam optimization over 15 epochs, with model performance evaluated using Mean Absolute Error (MAE), Root Mean Squared Error (RMSE), Huber loss, and Mean Squared Error (MSE metrics) on held-out test sets. Full implementation details are provided in Appendix~\ref{appendix:methodology}.

\section{Results}

An ablation study was conducted to assess the contribution of each component of the Temp-AI-Estimator (AI-Temp) model. We compared the full model against five simplified variants, each removing one key component: adversarial domain alignment (Adv), calibration (Cal), physical modulation ($F_{phy}$), external correction ($F_{ext}$), and finally, leaving only the LSTM backbone. Evaluation was performed on test sets from both Nigeria and The Gambia.

As shown in Table~\ref{tab:ablation}, the full AI-Temp model achieved the best performance on the Nigeria test set, with the lowest MAE (1.45°C), RMSE (1.75°C), scaled MSE (0.23), and Huber loss (0.226). Interestingly, in The Gambia, the best results across all metrics were observed when adversarial training was removed. This suggests that while domain adversarial training (via the GRL and discriminator) improves generalization from Tanzania to Nigeria by learning domain-invariant features, it may harm performance on structurally different data, such as Gambian houses, which lack windows and differ significantly from Tanzanian schools. In such cases, adversarial alignment may suppress critical physical cues needed for accurate prediction.

\begin{table}
  \caption{Ablation study results comparing model variants across Nigeria and Gambia test sets}
  \label{tab:ablation}
  \centering
  \begin{tabular}{lcccccccc}
    \toprule
    & \multicolumn{4}{c}{Nigeria} & \multicolumn{4}{c}{Gambia} \\
    \cmidrule(lr){2-5} \cmidrule(lr){6-9}
    Model & MAE & RMSE & MSE & Huber & MAE & RMSE & MSE & Huber \\
    \midrule
    AI-Temp (Full) & \textbf{1.45} & \textbf{1.75} & \textbf{0.47} & \textbf{0.23} & 0.72 & 0.93 & 0.13 & 0.067 \\
    AI-Temp (- Adv) & 1.70 & 2.05 & 0.65 & 0.31 & \textbf{0.65} & \textbf{0.84} & \textbf{0.11} & \textbf{0.054} \\
    AI-Temp (- Cal) & 4.51 & 4.97 & 3.80 & 1.30 & 2.13 & 2.69 & 1.12 & 0.46 \\
    AI-Temp (- $F_{phy}$) & 1.58 & 1.87 & 0.54 & 0.26 & 0.74 & 0.97 & 0.14 & 0.072 \\
    AI-Temp (- $F_{ext}$) & 1.77 & 2.08 & 0.67 & 0.32 & 0.77 & 0.99 & 0.15 & 0.075 \\
    AI-Temp (LSTM) & 1.75 & 2.12 & 0.69 & 0.32 & 3.10 & 3.97 & 2.42 & 0.82 \\
    \bottomrule
  \end{tabular}
\end{table}

\begin{figure}[htbp]
  \centering
  \includegraphics[width=0.9\linewidth]{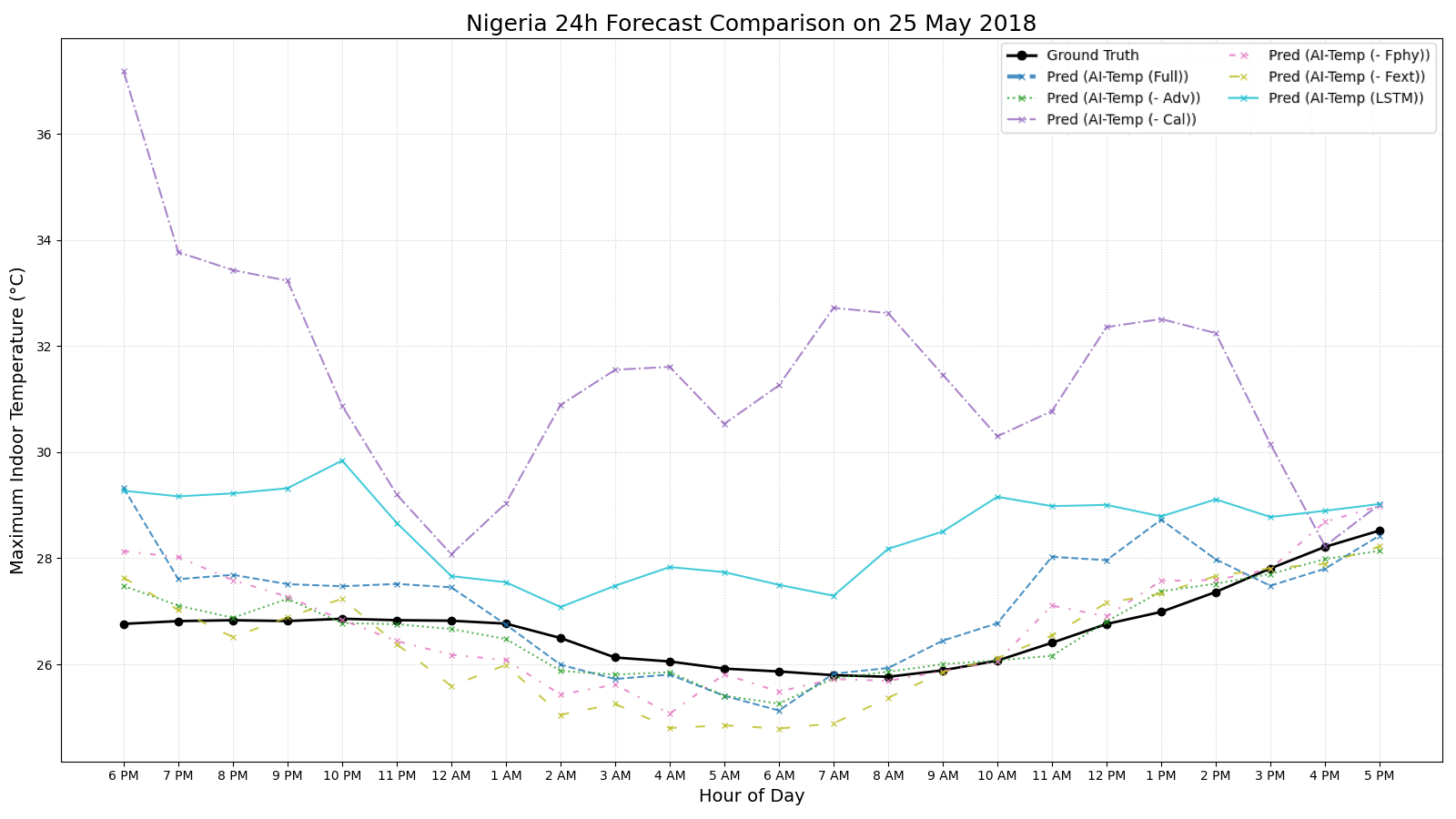}
  \caption{24 hour indoor temperature forecast in a Nigerian School.}
  \label{fig:nigeria}
\end{figure}

\begin{figure}[htbp]
  \centering
  \includegraphics[width=0.9\linewidth]{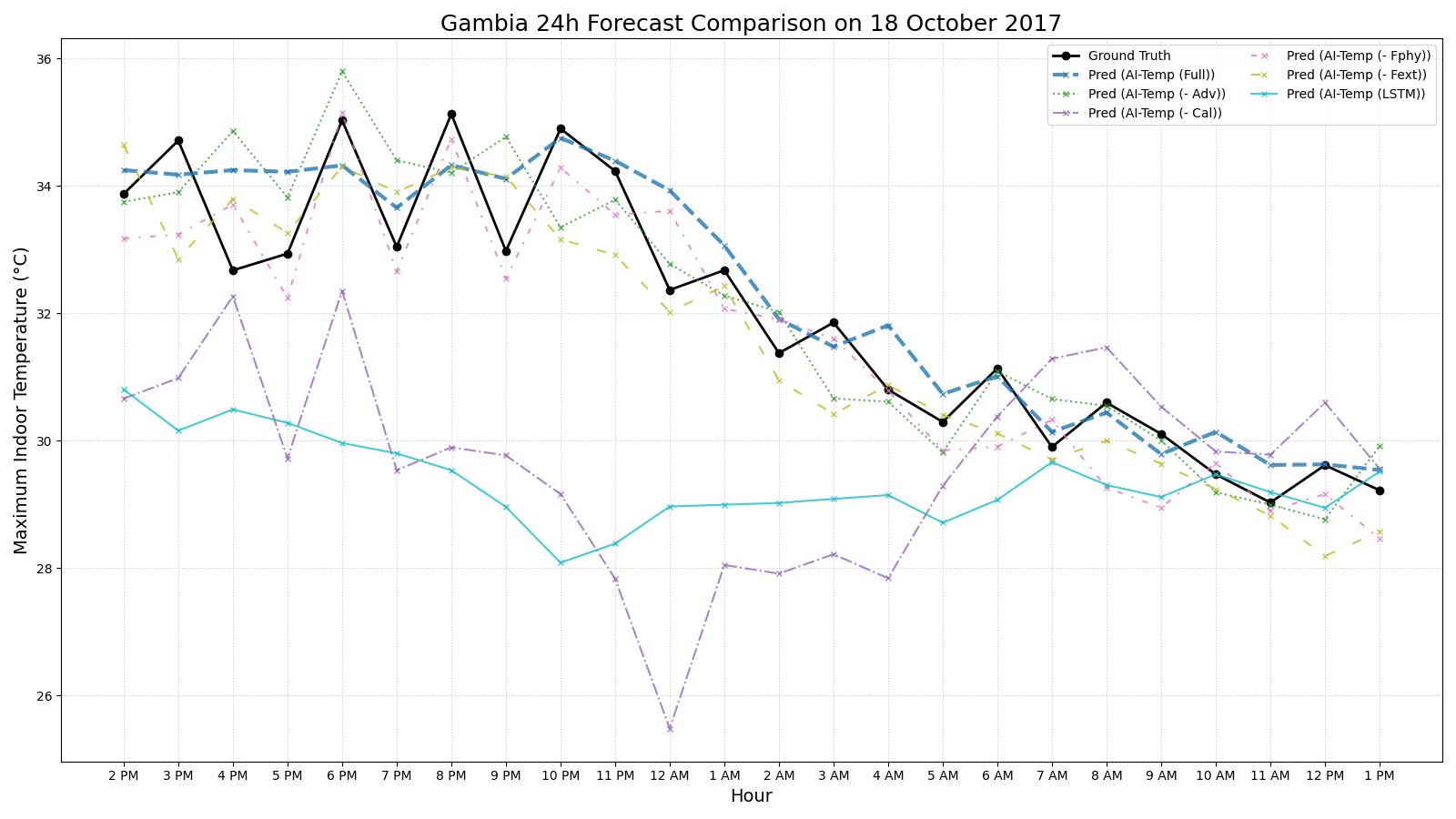}
  \caption{24 hour indoor temperature forecast in a Gambian House.}
  \label{fig:gambia}
\end{figure}

In both countries, removing calibration resulted in the largest accuracy drop, highlighting the importance of even limited labeled data in the target domain. Forecast visualizations (Figures  \ref{fig:nigeria} and  \ref{fig:gambia}) reinforce this finding: the full model closely follows observed temperatures in Nigeria, while the (-Adv) variant performs best in The Gambia. In contrast, models without calibration or with only the LSTM backbone show clear bias and reduced temporal stability.

\section{Discussions}
This study adapted the Temp-AI-Estimator model [3] for predicting indoor temperatures in naturally ventilated schools and homes, demonstrating promising cross-country generalization across three SSA countries. Several opportunities for future work emerge. First, incorporating occupancy schedules (e.g., class vs. break times) and evaluating the role of physical factors such as building orientation may improve prediction accuracy. Second, exploring more expressive model architectures, such as DEML [17], could enhance model performance. Although this study focused solely on indoor temperature as a proxy for thermal comfort, future models could incorporate relative humidity and dew point to estimate thermal comfort indices directly. There is also a need for longer-term, high-frequency datasets that capture detailed physical characteristics (e.g., ventilation quality, electricity usage, and structural features), essential to support model generalizability and real-world applicability.

\section{Conclusion}
This study introduced a domain-informed AI model for predicting indoor temperatures in naturally ventilated buildings across Sub-Saharan Africa, using only accessible weather and basic building data. The model demonstrated strong cross-country generalization with mean absolute errors of 1.45°C (Nigeria) and 0.65°C (Gambia), with ablation studies highlighting the critical role of calibration. This lightweight approach offers a practical solution for thermal comfort assessment in resource-constrained educational and residential settings, supporting evidence-based infrastructure planning and climate adaptation strategies.

\section*{References}

[1] Aguilera, J.J., Korsholm Andersen, R.\ \& Toftum, J.\ (2019) Prediction of indoor air temperature using weather data and simple building descriptors. {\it International Journal of Environmental Research and Public Health} 16(22):4349.

[2] Anderson, K., Hansen, C., Holmgren, W., Jensen, A., Mikofski, M.\ \& Driesse, A.\ (2023) pvlib python: 2023 project update. {\it Journal of Open Source Software} 8(92):5994.

[3] Bischof, R., Sprenger, M., Riedel, H., Bumann, M., Walczok, W., Drass, M.\ \& Kraus, M.A.\ (2023) Temp-AI-Estimator: Interior temperature prediction using domain-informed Deep Learning. {\it Energy and Buildings} 297:113425.

[4] Carrasco-Tenezaca, M., Jatta, E., Jawara, M.\ et al.\ (2021) Effect of roof colour on indoor temperature and human comfort levels, with implications for malaria control: a pilot study using experimental houses in rural Gambia. {\it Malaria Journal} 20:423.

[5] Clemente, A.V., Nocente, A.\ \& Ruocco, M.\ (2023) Global Transformer Architecture for Indoor Room Temperature Forecasting. In {\it Journal of Physics: Conference Series} 2600(2):022018. IOP Publishing.

[6] Copernicus Climate Change Service (C3S)\ (2023) ERA5 hourly data on single levels from 1940 to present. Copernicus Climate Change Service Climate Data Store (CDS).

[7] Jiang, L., Wang, X., Wang, L., Shao, M.\ \& Zhuang, L.\ (2021) A hybrid ANN-LSTM based model for indoor temperature prediction. In {\it 2021 IEEE 16th Conference on Industrial Electronics and Applications (ICIEA)}, pp.\ 1724--1728. IEEE.

[8] Laukkarinen, A.\ \& Vinha, J.\ (2024) Long-term prediction of hourly indoor air temperature using machine learning. {\it Energy and Buildings} 325:114972.

[9] Laterite and Fab Inc.\ (2024) Temperature, rainfall, and learning -- evidence from school surveys in Tanzania. Dar es Salaam, Tanzania.

[10] Mun, J.\ \& Park, C.S.\ (2025) Beyond correlation: A causality-driven model for indoor temperature control. {\it Energy and Buildings} 338:115739.

[11] Munonye, C.\ \& Maduabum, A.\ (2024) Classrooms and comfort temperature in South East Nigeria. {\it World Journal of Advanced Research and Reviews} 24(2):1564--1573.

[12] Mustapha, T.D., Hassan, A.S., Khozaei, F.\ \& Onubi, H.O.\ (2024) Examining thermal comfort levels and ASHRAE Standard-55 applicability: a case study of free-running classrooms in Abuja, Nigeria. {\it Indoor and Built Environment} 33(1):8--22.

[13] Pule, V., Mathee, A., Melariri, P., Kapwata, T., Abdelatif, N., Balakrishna, Y., Kunene, Z., Mogotsi, M., Wernecke, B.\ \& Wright, C.Y.\ (2021) Classroom temperature and learner absenteeism in public primary schools in the Eastern Cape, South Africa. {\it International Journal of Environmental Research and Public Health} 18(20):10700.

[14] Qi, D., Yi, X., Guo, C., Huang, Y., Zhang, J., Li, T.\ \& Zheng, Y.\ (2024) Spatio-temporal consistency enhanced differential network for interpretable indoor temperature prediction. In {\it Proceedings of the 30th ACM SIGKDD Conference on Knowledge Discovery and Data Mining}, pp.\ 5590--5601.

[15] Xing, T., Sun, K.\ \& Zhao, Q.\ (2023) MITP-Net: A deep-learning framework for short-term indoor temperature predictions in multi-zone buildings. {\it Building and Environment} 239:110388.

[16] Xu, C., Chen, H., Wang, J., Guo, Y.\ \& Yuan, Y.\ (2019) Improving prediction performance for indoor temperature in public buildings based on a novel deep learning method. {\it Building and Environment} 148:128--135.

[17] Yu, W., Nakisa, B., Ali, E., Loke, S.W., Stevanovic, S.\ \& Guo, Y.\ (2023) Sensor-based indoor air temperature prediction using deep ensemble machine learning: An Australian urban environment case study. {\it Urban Climate} 51:101599.

[18] Zippenfenig, P.\ (2023) Open-Meteo.com Weather API [Computer software]. Zenodo.

\section{Appendix}
\label{appendix:methodology}
\appendix

\begin{figure}[htbp]
  \centering
  \includegraphics[width=0.8\linewidth]{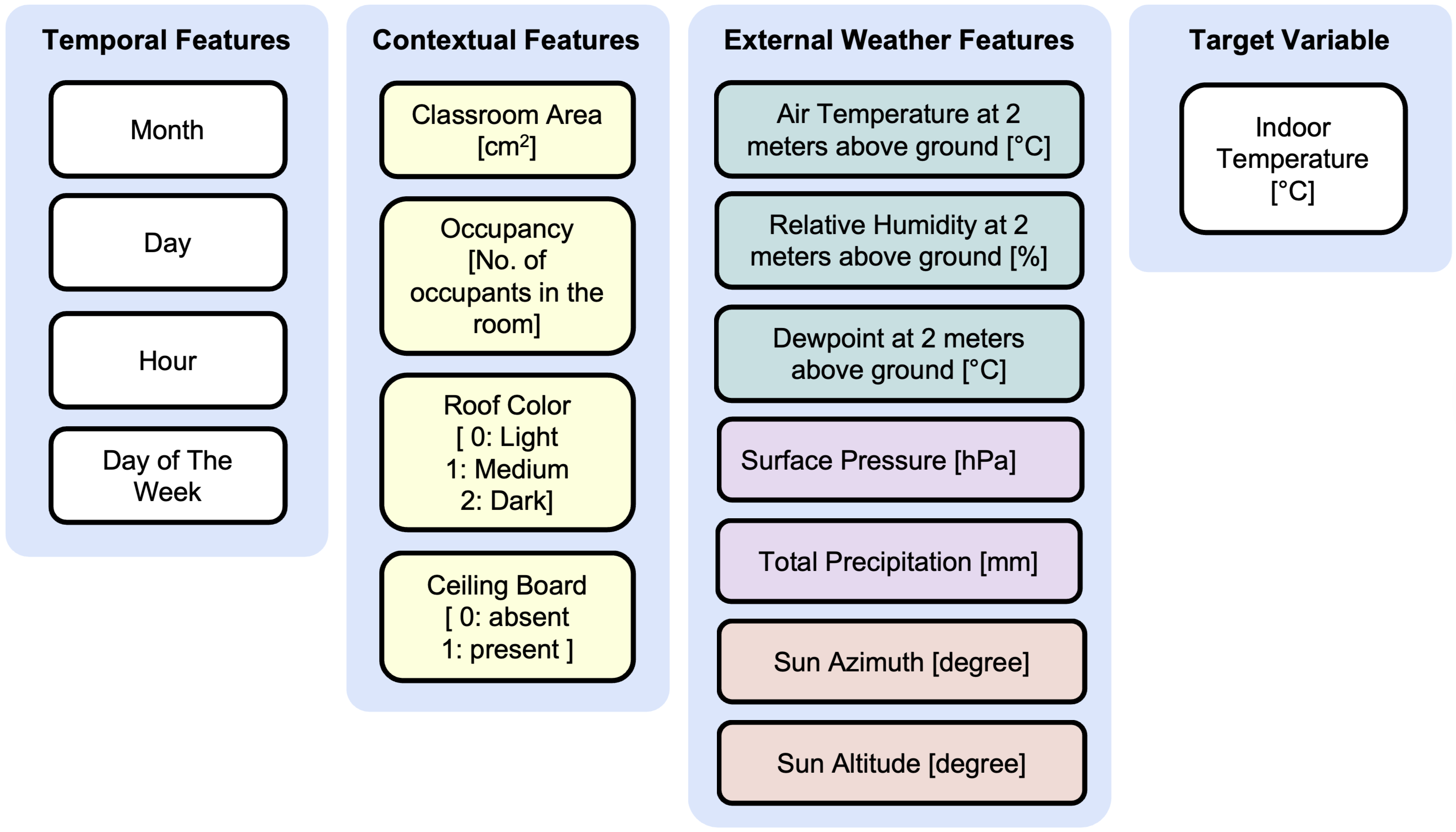}
  \caption{Categorized feature set used in the AI model.}
  \label{fig:featureset}
\end{figure}

\section{Methodology Details}
\label{appendix:methodology}

\subsection{Data Preprocessing and Feature Engineering}
\label{appendix:data-preprocessing}

Raw data from Tanzania, The Gambia, and Nigeria datasets were initially recorded at minute-level resolution. To align with hourly external weather data, aggregation to hourly intervals was performed, retaining hourly maximum temperatures to capture critical periods when indoor temperatures surpass thermal comfort thresholds.

Feature engineering involved several key steps:
\begin{itemize}
\item Converting categorical variables (roof color, ceiling board presence) into numerical codes
\item Encoding temporal features and sun azimuth/elevation as cyclical sine-cosine pairs
\item Scaling numeric features (outdoor temperature, humidity, precipitation) using StandardScaler\footnote{\url{https://scikit-learn.org/stable/modules/generated/sklearn.preprocessing.StandardScaler.html}}
\end{itemize}

\subsection{Sliding Window Configuration and Domain Splits}
\label{appendix:sliding-window}

The processed data were structured into sliding windows to provide historical context for predicting indoor temperatures. A look-back window size of 12 hours and prediction horizon of 24 hours were used, creating input tensors $X \in \mathbb{R}^{N \times W \times F}$ and output tensors $Y \in \mathbb{R}^{N \times H}$, where $N$ is the number of samples, $W$ = 12 hours, $H$ = 24 hours, and $F$ is the feature dimension.

To prepare for domain adaptation, the Nigeria and Gambia datasets (target domains) were chronologically split:
\begin{itemize}
\item \textbf{Adaptation Set:} First 90\% of samples
\begin{itemize}
\item \textbf{Unsupervised Domain Subset:} First 90\% of adaptation samples (no labels)
\item \textbf{Calibration Subset (Cal):} Last 10\% of adaptation samples (with labels)
\end{itemize}
\item \textbf{Test Set:} Final 10\% of samples reserved for evaluation
\end{itemize}

\subsection{Domain-Informed Model Architecture}
\label{appendix:architecture}
The adapted Temp-AI-Estimator architecture [3] consists of four modules integrated into a single framework:

\begin{itemize}
\item \textbf{LSTM Backbone (LSTM):} Captures temporal dependencies using differences of feature values to ensure stationarity. Outputs base temperature predictions.

\item \textbf{External-Correction Branch ($\mathbf{F}_{\mathbf{ext}}$):} A fully-connected Multi-Layer Perceptron (MLP) operating on the last step's feature values, predicting a scalar adjustment ($\delta_{ext}$) that corrects cumulative predictions from the LSTM.

\item \textbf{Physical-Modulation Branch ($\mathbf{F}_{\mathbf{phy}}$):} Another MLP that processes physical characteristics (roof color, ceiling board presence, occupancy, classroom area). It outputs two parameters, scaling ($s_{c}$) and shifting ($s_{h}$) factors that modulate the base predictions to account for building-specific conditions.

\item \textbf{Domain Discriminator (Adv):} Implements adversarial training via a Gradient Reversal Layer (GRL). This encourages domain-invariant feature extraction, facilitating better generalization across countries.
\end{itemize}

The final predicted temperature ($\widehat{y}$) combines these components:

\begin{equation}
\widehat{y} = \left( \widehat{y_{\text{base}}} \cdot s_{c} \right) + \delta_{\text{ext}} + s_{h}
\end{equation}

\subsection{Training Procedure}
\label{subsec:training}

The model was trained using three primary loss functions:

\begin{itemize}
\item \textbf{Supervised Regression Loss:} Huber loss calculated on labeled source (Tanzania) data and labeled calibration data (Nigeria/Gambia).

\item \textbf{Adversarial Domain Loss:} Binary cross-entropy (BCE) loss from the domain discriminator, scaled via GRL, promoting domain invariance.

\item \textbf{Total Loss:} Weighted combination:
\end{itemize}

\begin{equation}
L_{\text{total}} = L_{\text{Hub, source}} + w_{\text{cal}} \cdot L_{\text{Hub, calibration}} + 0.1 \cdot L_{\text{BCE, domain}}
\end{equation}

\subsubsection{Training Configuration}
\begin{itemize}
\item Optimizer: Adam with learning rate = $10^{-3}$
\item Batch size: 64
\item Epochs: 15
\item GRL adversarial coefficient ($\lambda_{\text{adv}}$): Gradually increased from 0 to 0.01 following a sigmoid schedule after initial warm-up period
\end{itemize}

\subsection{Evaluation Metrics}
\label{appendix:evaluation}

Model performance was evaluated on held-out test datasets from Nigeria and The Gambia using:
\begin{itemize}
\item Mean Absolute Error (MAE)
\item Root Mean Squared Error (RMSE) 
\item Mean Huber Loss
\item Mean Squared Error (MSE)
\end{itemize}

The latter two metrics were computed on scaled data for consistency with training procedures.

\end{document}